%% file: sample-sigconf.tex
\documentclass[sigconf]{acmart}

\usepackage{placeins}

\usepackage{tikz} 
\usepackage{listings}
\usepackage{lscape}
\usepackage{minted}
\usepackage{dirtree}
\usepackage{float}

\usepackage{enumitem}
\usepackage{graphicx}
\usepackage{multicol, latexsym}
\usepackage{blindtext}
\usepackage[skip=5pt,font=small]{caption}
\usepackage{longtable}
\usepackage{csquotes}
\usepackage{amsfonts}
\usepackage{amsmath}
\usepackage{amsthm}
\usepackage{algorithm}
\usepackage{tabularx}
\usepackage{capt-of,lipsum} 
\usepackage{subfigure} 
\usepackage{subcaption} 

\usepackage{listings}
\usepackage{comment}
\usepackage{diagbox}
\usepackage{pdflscape}
\usepackage{booktabs}
\usepackage{makecell}
\usepackage{balance}
\usepackage{listings}
\usepackage{xcolor}
\usepackage{array}

\usepackage{mathtools, nccmath}

\usepackage{siunitx}
\DeclareSIUnit[number-unit-product = { }] \dBm{dBm}

\usepackage{algpseudocode}
\usepackage{algorithm}

\usepackage{listing-styles}

\algnewcommand\algorithmicforeach{\textbf{for each}}
\algdef{S}[FOR]{ForEach}[1]{\algorithmicforeach\ #1\ \algorithmicdo}

\definecolor{myblue}{rgb}{0.09,0.20,0.34}
\definecolor{mygreen}{rgb}{0,0.6,0}
\definecolor{mygray}{rgb}{0.98,0.98,0.98}
\definecolor{myorange}{rgb}{0.92,0.49,0.34}
\definecolor{mywhite}{rgb}{1.0,1.0,1.0}

\usepackage[scaled]{helvet}
\usepackage[T1]{fontenc}

\AtBeginDocument{%
  }

\usepackage{pifont}            
\usepackage{wasysym}
\newcommand*\feature[1]{\ifcase#1 -\or\LEFTcircle\or\CIRCLE\fi}
\usepackage{threeparttable}

\begin{document}


\title{From Noise to Knowledge: A Comparative Study of Acoustic Anomaly Detection Models in Pumped-storage Hydropower Plants}



\author{Karim Khamaisi}
\affiliation{%
  \institution{University of St. Gallen}
  \city{Dornbirn}
  \country{Austria}}
\email{karim.khamaisi@unisg.ch}

\author{Nicolas Keller}
\affiliation{%
  \institution{University of St. Gallen}
  \city{St. Gallen}
  \country{Switzerland}}
\email{nicolas.keller@student.unisg.ch}

\author{Stefan Krummenacher}
\affiliation{%
  \institution{University of St. Gallen}
  \city{St. Gallen}
  \country{Switzerland}}
\email{stefan.krummenacher@student.unisg.ch}

\author{Valentin Huber}
\affiliation{%
  \institution{University of St. Gallen}
  \city{St. Gallen}
  \country{Switzerland}}
\email{valentin.huber@student.unisg.ch}

\author{Bernhard Fässler}
\affiliation{%
  \institution{illwerke vkw AG}
  \city{Bregenz}
  \country{Austria}}
\email{bernhard.faessler@illwerkevkw.at}

\author{Bruno Rodrigues}
\affiliation{%
  \institution{University of St. Gallen}
  \city{Dornbirn}
  \country{Austria}}
\email{bruno.rodrigues@unisg.ch}

\renewcommand{\shortauthors}{Khamaisi et al.}
\begin{abstract}

In the context of industrial factories and energy producers, unplanned outages are highly costly and difficult to service. However, existing acoustic-anomaly detection studies largely rely on generic industrial or synthetic datasets, with few focused on hydropower plants due to limited access. This paper presents a comparative analysis of acoustic-based anomaly detection methods, as a way to improve predictive maintenance in hydropower plants. We address key challenges in the acoustic preprocessing under highly noisy conditions before extracting time- and frequency-domain features. Then, we benchmark three machine learning models: LSTM AE, K-Means, and OC-SVM, which are tested on two real-world datasets from the Rodundwerk II pumped-storage plant in Austria, one with induced anomalies and one with real-world conditions. The One-Class SVM achieved the best trade-off of accuracy (ROC AUC 0.966–0.998) and minimal training time, while the LSTM autoencoder delivered strong detection (ROC AUC 0.889–0.997) at the expense of higher computational cost.


\end{abstract}




\maketitle

\input{sections/1_introduction}

\input{sections/2_background}

\input{sections/3_design}

\input{sections/4_results}

\input{sections/6_conclusions}

\begin{acks}
This work was conducted in the context of a joint project with illwerke vkw AG, Bregenz, Austria, to whom the authors express gratitude for all support and discussions on technical aspects, and access to the Rodundwerk II hydropower plant.
\end{acks}

\bibliographystyle{ACM-Reference-Format}
\bibliography{bib/sample-base}

\appendix

\end{document}

%% file: sections/1_introduction.tex
\section{Introduction} \label{chp:introduction}

Maintenance is a critical function in industrial environments due to the significant operational and financial impacts associated with equipment failure. Recent estimates highlight that unplanned downtime costs industries billions annually, with individual facilities losing millions per event \cite{siemens2023downtime}. Such unexpected interruptions are particularly consequential in energy production facilities like pumped-storage hydropower plants, where operational continuity is essential to maintain grid stability and minimize economic impacts.

Hydropower plant equipment, including turbines and generators, is central to plant operations. Failures in these mechanical parts can result in costly downtime. Downtime costs are highly variable, primarily depending on fluctuating market energy prices. For instance, the Rodundwerk II pumped-storage (\textit{cf.} Section \ref{sec:relatedwork}) hydropower plant in Vorarlberg, Austria, has a full-load capacity of 295 MW, translating to 295 MWh of electricity production per hour. With a wholesale price of 97.30 Euro/MWh \cite{EControl.} (as of February, 2025), downtime can cost over 28,000 Euros per hour \cite{illwerkeRodundwerk}. Moreover, since pumped-storage hydropower plants typically function as peak-load power plants, selling electricity predominantly at times of high market prices, actual downtime costs can significantly exceed this estimate.

Predictive maintenance (PdM) proactively anticipates equipment failures, enabling maintenance timeing that avoids both unnecessary downtime and catastrophic failures \cite{Mobley2002, ZONTA2020106889}. Among various PdM approaches, acoustic-based anomaly detection stands out for its non-intrusive nature, cost-effectiveness, and sensitivity to early-stage mechanical defects \cite{diFiore2022, konig2021, Khanjari2024}. Acoustic signals carry vital diagnostic information, reflecting subtle deviations in machinery operations that precede potential failures. However, acoustic-anomaly detection methods often face challenges in industrial environments characterized by high noise levels, complex acoustic patterns, and limited labeled anomaly data \cite{loScudo2023, tagawa2021}.

\input{tables/relatedwork}

Although acoustic anomaly detection has been extensively studied, existing literature predominantly relies on generic industrial or synthetic datasets \cite{purohit2019mimii, duman2020, diFiore2022, Bayram2021}, with limited representation from actual hydropower plant operations. This gap exists mostly due to restricted access to critical infrastructure, such as pumped-storage hydropower facilities, whose acoustic characteristics significantly differ from standard industrial machinery. Consequently, existing models \cite{konig2021,loScudo2023, tagawa2021} trained on generic data often fail to generalize effectively to the acoustic conditions and anomalies characteristic of hydropower plants.

We provide systematic comparative analysis of acoustic anomaly detection methods explicitly applied to a pumped-storage hydropower plant, Rodundwerk II in Austria managed and operated by illwerke vkw AG \cite{illwerkeRodundwerk}. \textbf{Contributions} are summarized as follows:
\begin{itemize}
    \item We present an acoustic preprocessing pipeline tailored to mitigate noise and handle the challenging acoustic environment of hydropower plants.
    \item Benchmark three prominent machine learning models: LSTM Autoencoder (LSTM AE), K-Means clustering, and One-Class SVM (OC-SVM), using two distinct real-world datasets acquired from Rodundwerk II: one with induced anomalies under controlled conditions, and another featuring naturally occurring operational anomalies.
    \item We discuss strengths and weaknesses of each model, providing guidelines for practitioners on selecting appropriate methods based on operational constraints.
\end{itemize}

The remainder of this paper is organized as follows:
Section~\ref{sec:relatedwork} reviews existing literature on acoustic anomaly detection, focusing on both traditional signal processing and machine learning approaches.  
Section~\ref{sec:design} presents the proposed framework, detailing its preprocessing pipeline and model architectures.  
Section~\ref{sec:results} discusses experimental results and provides a comparative evaluation real-world datasets.  
Finally, Section~\ref{sec:conclusion} offers concluding remarks and outlines directions for future research.

%% file: tables/relatedwork.tex
\begin{table*}[!htpb]
  \caption{Selected Related Work on Acoustic Anomaly-detection}
  \label{tab:comparative_analysis}
  \resizebox{\textwidth}{!}{%
    \begin{tabular}{l l l l l}
      \toprule
      \textbf{Study} &
      \textbf{Domain / Data} &
      \textbf{Feature Rep.} &
      \textbf{Model(s)} &
      \textbf{Key Insight} \\
      \midrule
      Müller et al. (2020)\,\cite{muller2020acoustic} &
      Factory machinery &
      MelSpec\,+\,CNN feats &
      IF, GMM, OC-SVM, AE &
      CNN features beat CAE in noise \\

      Meire et al. (2019)\,\cite{meire2019} &
      Industrial bench setup &
      MFCC, MelSpec &
      OC-SVM, AE &
      HW-friendly AEs for edge devices \\

      Bayram et al. (2021)\,\cite{Bayram2021} &
      Chemical process plant &
      MelSpec &
      Conv-LSTM-AE, CAE &
      Sequential AEs allow real-time \\

      Ferraro et al (2023)\,\cite{Ferraro2023UnsupervisedAnomaly} &
      Multi-factory lines &
      MelSpec &
      LSTM-AE, CNN-AE &
      Sliding-window edge inference \\

      Coelho et al. (2022)\,\cite{Coelho2022DeepAutoencoders} &
      Factory \& vehicle &
      MFEC, MelSpec &
      Dense/CNN/LSTM AEs &
      Depth vs.\ noise-robustness study \\

      Duman et al. (2020)\,\cite{duman2020} &
      Industrial plant &
      MelSpec &
      CAE &
      Lightweight CAE for on-site use \\

      Ahn et al. (2021)\,\cite{Ahn2021} &
      Vehicle acoustics &
      Raw audio &
      SVM, K-Means, CNN &
      Multi-mic array boosts recall \\

      Tagawa et al. (2021)\,\cite{tagawa2021} &
      Rotating machinery &
      Spectral coeff. &
      SVM, Auto-corr. &
      Spectral coeff.\ key for bearings \\

      Purohit et al. (2019)\,\cite{purohit2019mimii} &
      MIMII (public) &
      Log-MelSpec &
      VAE, AE &
      Widely used open benchmark \\
      \bottomrule
    \end{tabular}}
\end{table*}

%% file: sections/2_background.tex
\section{Background and Related Work} \label{sec:relatedwork}

\textbf{Pumped storage hydropower plants (PSHP).} Traditional hydropower plants generate electricity using water stored behind a dam, flowing downstream to drive turbines \cite{kishor2007review}. In contrast, PSHPs utilize mountainous topography with two water reservoirs \cite{perez2015trends}. During low demand, water is pumped from the lower reservoir to the upper reservoir using surplus energy; during high demand, stored water is released to generate electricity. 

Unlike traditional hydropower plants, PSHPs require turbines to operate in dual mode (generation and pumping), leading to frequent operational changes that increase stresses, vibrations, and dynamic loads. These factors accelerate turbine fatigue and wear, causing issues like pump cavitation \cite{zhao2025comprehensive}. Anomaly detection facilitates early fault detection to prevent costly downtime \cite{diFiore2022, Khanjari2024}, with acoustic-based methods leveraging non-intrusive signals to capture subtle mechanical deviations prior to critical failures \cite{loScudo2023, tagawa2021}. 

\textbf{Feature extraction techniques.} Robust feature extraction is essential for accurately representing acoustic signals. Traditional methods like Short-Time Fourier Transform (STFT) and Mel-scale transformations convert raw audio signals into time-frequency spectrograms \cite{Bayram2021, ZONTA2020106889}. Mel-spectrograms are widely adopted due to their perceptual alignment with human auditory systems \cite{meire2019}, while Mel-Frequency Cepstral Coefficients (MFCCs) capture critical temporal and spectral characteristics that enhance anomaly detection \cite{Bayram2021, Coelho2022DeepAutoencoders}.

\textbf{Machine learning models.} Acoustic anomaly detection typically employs machine learning (ML) methods, broadly categorized into unsupervised and deep learning approaches. Density-based models like Local Outlier Factor (LOF)identify anomalies based on local data densities without extensive labeled data \cite{Breunig2000, Wang2022}. Clustering methods like K-Means offer computational simplicity but struggle with overlapping acoustic patterns \cite{Ahn2021}. Deep learning methods like Autoencoders (AEs) detect anomalies through reconstruction errors, proving effective in complex environments \cite{duman2020, Ferraro2023UnsupervisedAnomaly}, though they require substantial computational resources often unavailable in hydropower plants.

\begin{figure*}[h]
    \centering
        \subfigure[]{\includegraphics[width=.814\columnwidth, trim={6cm 0 6cm 0}, clip]{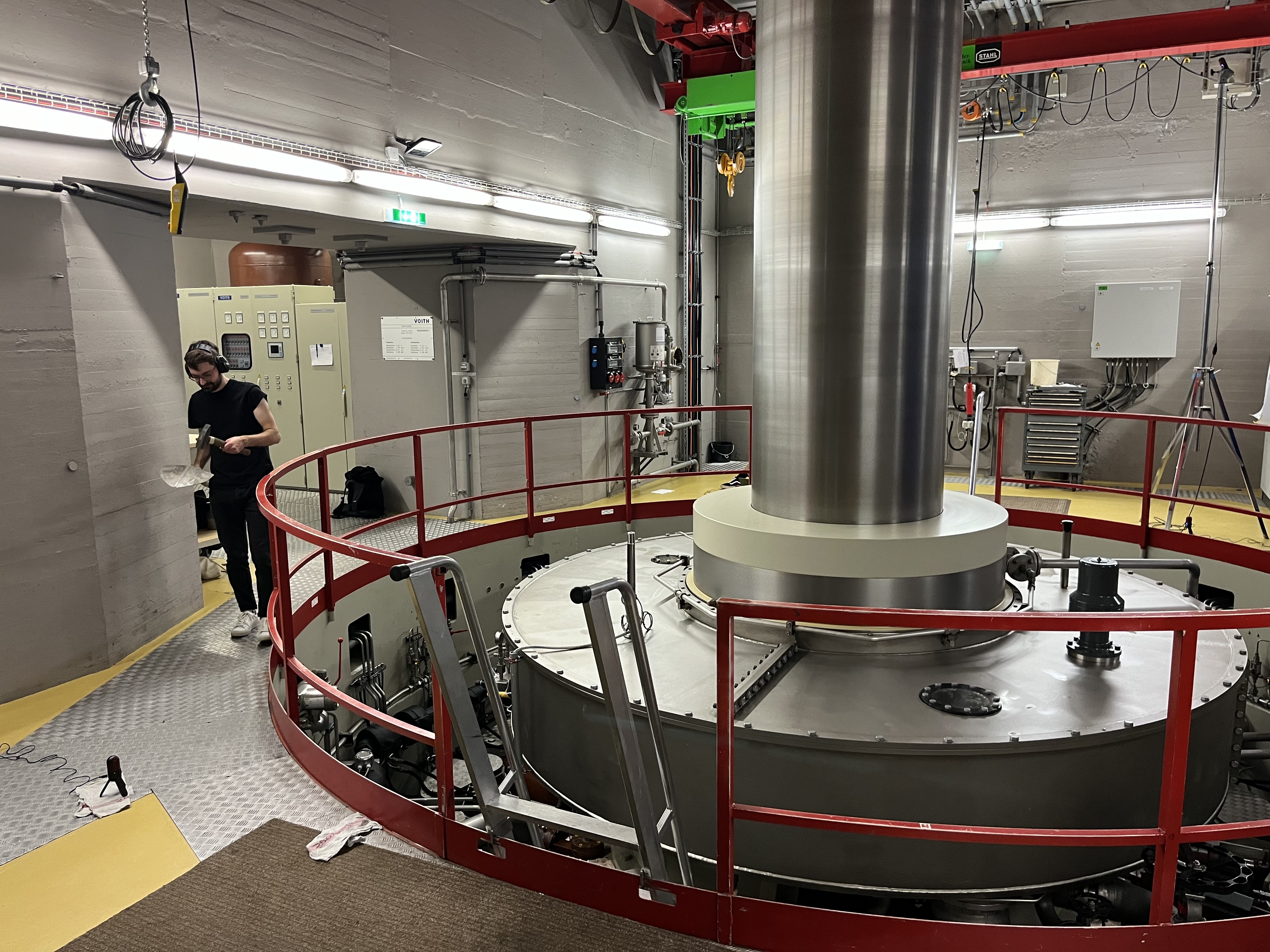}}  
        \subfigure[]{\includegraphics[width=.5\columnwidth]{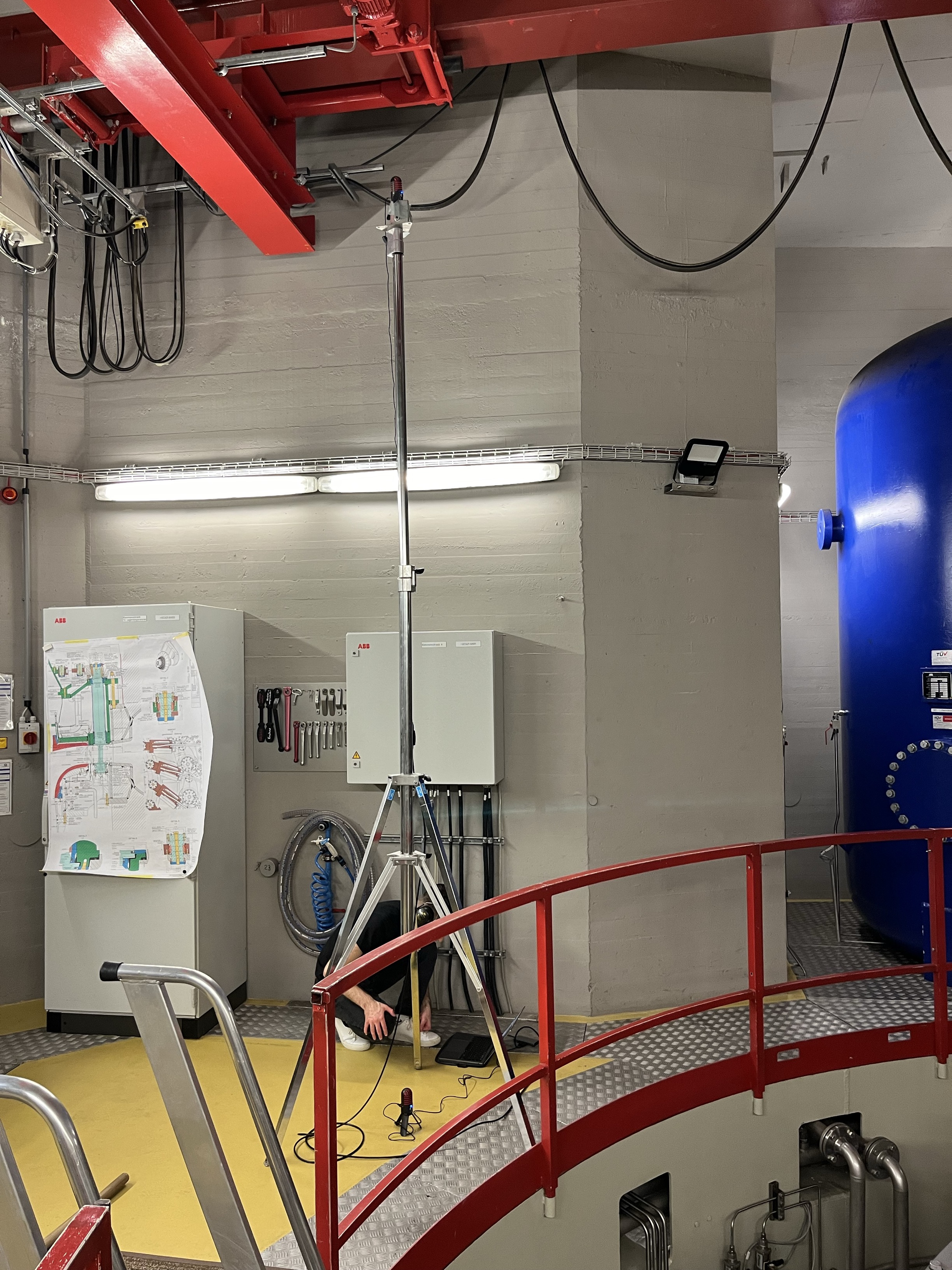}} 
        \subfigure[]{\includegraphics[width=.5\columnwidth,]{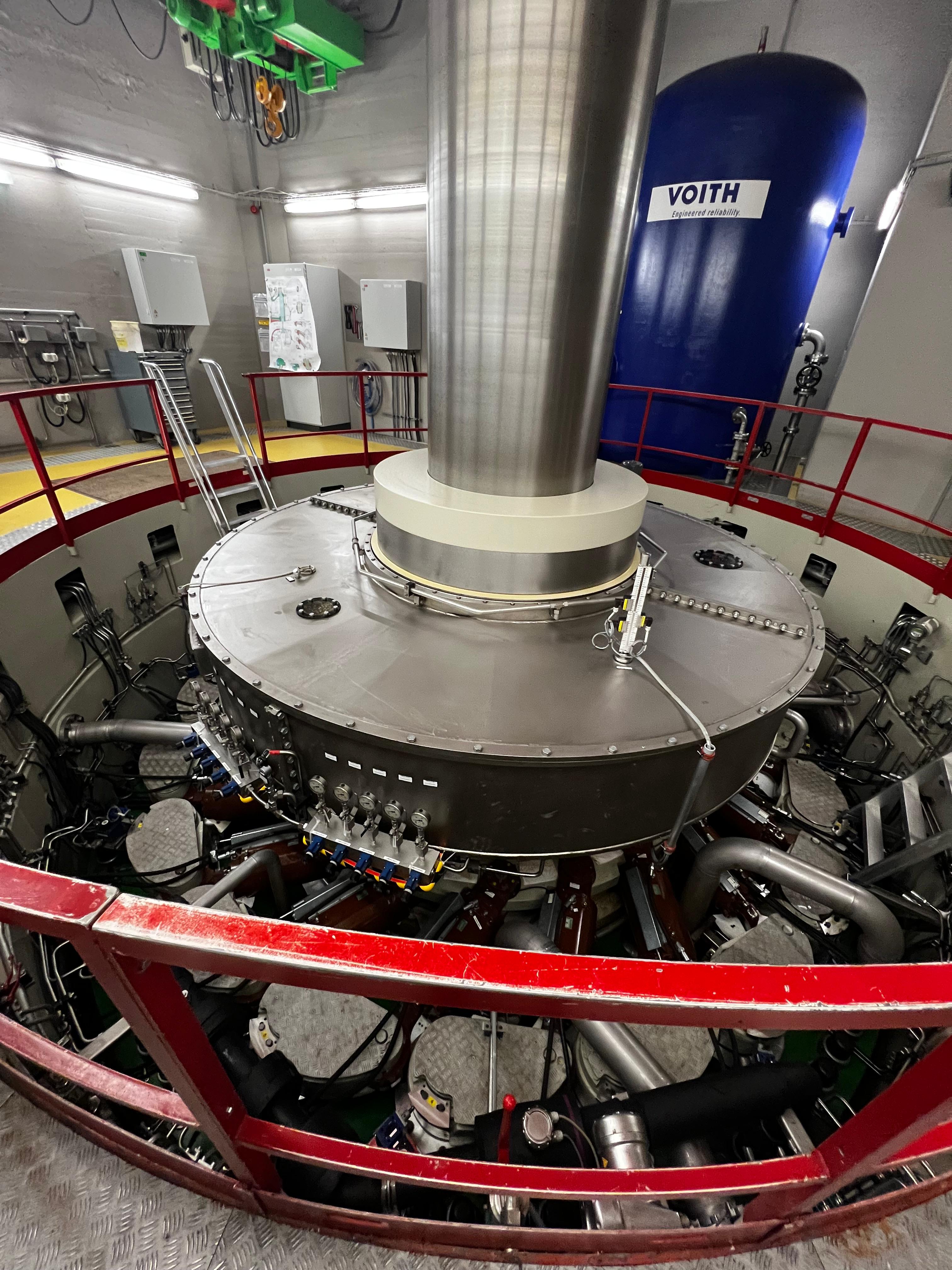}}
        \caption{Data Acquisition in PSHP Rodundwerk II: (a) audio recordings with induced anomalies (hammer and a shovel). (b) microphone stands in 4 corners capturing airborne audio. (c) shaft of the reversible francis-pump turbine; microphone stands were installed in each corner of the room.}
        \label{fig:deployment}
\end{figure*}

Table~\ref{tab:comparative_analysis} provides a concise comparative overview of recent relevant works, highlighting feature extraction methods, ML models used, and key objectives. Existing work relies mainly on generic or synthetic datasets, with few incorporating real-world data from specific applications such as hydroelectric plants \cite{purohit2019mimii, Bayram2021}. In terms of feature extraction, Mel-spectrograms are the most commonly used technique \cite{muller2020acoustic,meire2019, Bayram2021,Ferraro2023UnsupervisedAnomaly,Coelho2022DeepAutoencoders,duman2020}, followed by MFCCs and STFT \cite{meire2019}. This highlights the importance and effectiveness of converting raw audio signals into meaningful inputs for ML models.

The most used ML models are AEs, which are highly adaptable to specific audio signals and effectively detect abnormal sounds deviating from learned baselines. Various AE architectures appear across studies: \cite{Coelho2022DeepAutoencoders} explores three different architectures, \cite{Ferraro2023UnsupervisedAnomaly} compares LSTM-AE and CNN-AE models, while \cite{Bayram2021} utilizes Conv-LSTMAE and CAE. This diversity highlights standardization limitations, as no single model consistently outperforms others across all scenarios.
%

%% file: sections/3_design.tex
\section{Proposed Method}  
\label{sec:design}

\begin{figure}[b] 
  \centering
  \includegraphics[width=0.9\linewidth, trim={.5cm .5cm .5cm .5cm}, clip]{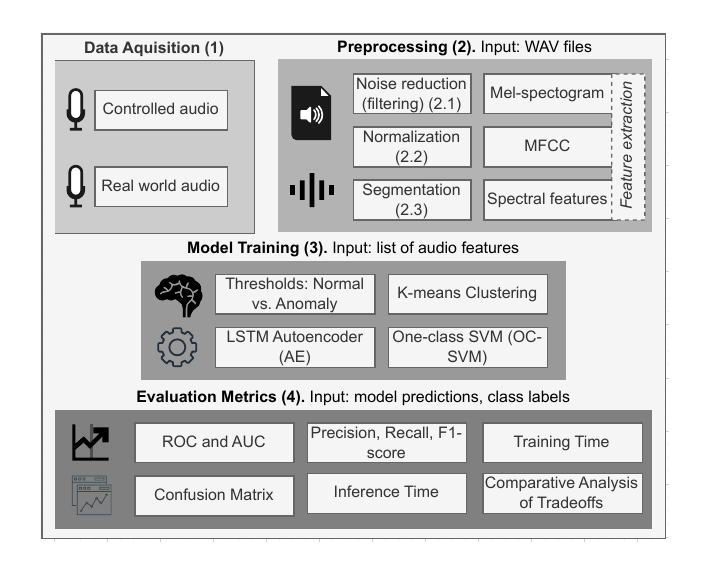}
  \caption{Overview of the acoustic anomaly detection framework}
  \label{fig:pipeline}
\end{figure}

Figure~\ref{fig:pipeline} illustrates the proposed acoustic anomaly detection pipeline. The pipeline addresses the challenges inherent in handling acoustic data from PSHP hydropower plants. It consists of four main stages, which are detailed in next subsections: Data Acquisition, Preprocessing, Model Benchmarking, and Evaluation Metrics.

The design of this pipeline reflects key considerations for handling real-world industrial acoustic-data, including noise resilience, efficient anomaly representation, and scalable deployment. The pipeline comprises four main stages:
\begin{enumerate}
    \item \textbf{Data Acquisition}: audio recordings of machine operations in controlled and real-world environments.
    \item \textbf{Preprocessing}: transformation of raw audio into normalized Mel-spectrograms.
    \item \textbf{Model Training}: benchmarking and evaluation of five selected models.
    \item \textbf{Evaluation Metrics}: identification of anomalies using reconstruction error and comparing with the benchmarks.
\end{enumerate}

\subsection{Data Acquisition}
Two distinct datasets were produced on-site to benchmark models for the acoustic anomaly detection. This was achieved by recording the machinery in real operating modes and with induced anomalies, splitting them into separate \textit{.wav} files, used for data preprocessing. 

The dataset with \textbf{induced anomalies} was recorded on-site to help in the first stage of fine-tuning models (cf Figure \ref{fig:deployment} - a). We used a hammer striking a shovel to resemble real‐world structural issues yet remain safe and repeatable. These synthetic knocks allow replication of realistic fault phenomena and thus form the basis for testing the framework under near–real‐world conditions while preserving consistency in labeling.

The \textbf{real dataset} was collected from an operational pumped storage hydropower plant under strict operational constraints that inherently limit data collection scope (in which the authors are thankful to the Rodundwerk's technical team). Access to such facilities is restricted due to safety regulations, security protocols, and the need for coordination with plant operators. 

\subsection{Preprocessing}

Prior to modelling, raw audio signals must be converted into representations that are both stable and discriminative in the context of acoustic anomaly detection. We use a Short-Time Fourier Transform (STFT) followed by a Mel-spectrogram conversion and a frame-based segmentation procedure. These steps are essential to transform the original time domain signals into a form more suited to machine learning methods. In addition, we used a combination of \textbf{noise reduction} to provide cleaner audio features and reduce the influence of noisy backgrounds; \textbf{normalization} using root mean square through average amplitude levels to ensure consistency while preserving feature stability; and \textbf{segmentation} splitting audio recordings into smaller, overlapping frames before being input to the ML model, facilitating feature extraction and maintaining balanced datasets \cite{sciencedirect_audio_segmentation}.

To extract audio features, we first apply a standard discrete \textit{Short-Time Fourier Transform (STFT)} \cite{1163359, mcfee2015librosa, cohen1995time} from the \textit{librosa} library with a 1024-sample window and a hop length of 512. By decomposing the audio into short, overlapping time segments, we obtain local frequency-domain information, enabling the detection of short-duration events (\textit{e.g.,} knocks or impacts). This time-frequency representation is critical for capturing the subtle temporal patterns that characterize anomalies in industrial or mechanical systems. Mathematically, the STFT is defined in Equation~\ref{eq:stft}, where we sum over windowed signal frames and apply the discrete Fourier transform.
\begin{equation} \label{eq:stft}
\mathrm{STFT}(n, k) 
= \sum_{m=0}^{L-1} 
    x(n + m)\, w(m)\, 
    e^{-j\,2\pi\,\frac{k}{N}\, m}
\end{equation}

From the STFT magnitude, we derive a 128‐band \textit{Mel‐spectrogram} \cite{1163420, mcfee2015librosa, malcolm1998auditory}. The Mel scale aligns more closely with human auditory perception, ensuring that frequency regions vital to detecting anomalies (\textit{e.g}., sudden high‐frequency bursts) are not overshadowed by less relevant areas of the spectrum. By compressing the frequency axis in this perceptual manner, the model is better able to focus on critical frequency components that differentiate normal from abnormal operations.  

\begin{equation} \label{eq:mel}
\mathrm{M}(m,n) 
  = \sum_{k=0}^{K-1} 
      H_{m}(k)\, |STFT(n, k)|^{2}
\end{equation}

In Equation~\ref{eq:mel}, each Mel bin is computed by summing the STFT power according to a triangular filter bank \(H_m\).

\begin{equation} \label{eq:mel_to_meldb}
\mathrm{M_{DB}}(m,n) 
\;=\; 
10 \,\log_{10}\!
\Bigl(
  \tfrac{M(m,n)}{\mathrm{max_{m, n}{(M(m,n ))}}}
\Bigr)
\end{equation}

The Mel-spectrogram is converted to decibels using Equation~\ref{eq:mel_to_meldb}, normalizing each cell by the global maximum and applying \(10 \log_{10}\). Then, each Mel‐spectrogram is min‐max normalized \cite{9214160} to a \([0,1]\) range as shown in Equation~\ref{eq:normalization}, which subtracts the minimum and divides by the overall range: 
\begin{equation} \label{eq:normalization}
x_{\mathrm{norm}} 
= \frac{x - \min(X)}{\max(X) - \min(X)}
\end{equation}

This standardization step makes training more robust by reducing amplitude-related variability and ensuring uniform feature scales across the dataset. As a result, the model is not biased by unusually loud or soft recordings, improving generalization to new, unseen data.

\begin{minipage}{0.95\linewidth} 
\begin{lstlisting}[caption={Code to Generate and Normalize Mel-Spectrograms},language=Python, label={lst:mel_spectrogram}]
time_per_frame = 0.6   
hop_ratio      = 0.2   
hop_length     = 512   

def generate_mel_spectrogram(audio_path):
    audio, sr = librosa.load(audio_path, sr=None)
    stft = librosa.stft(audio, n_fft=1024, hop_length=hop_length)
    mel  = librosa.feature.melspectrogram(S=np.abs(stft)**2, sr=sr, n_mels=128)
    mel_db = librosa.power_to_db(mel, ref=np.max)
    mel_db_norm = (mel_db - mel_db.min()) / (mel_db.max() - mel_db.min())
    return mel_db_norm, sr
\end{lstlisting}

\end{minipage}

The Python code in Listing~\ref{lst:mel_spectrogram} illustrates how the first steps are implemented in practice, including loading the audio data, applying an STFT, generating and normalizing Mel‐spectrograms. After computing the Mel‐spectrogram, we segment each spectrogram into overlapping frames. In our setup:
\begin{itemize}
    \item{Windowing:} A frame length of 0.512\,s is chosen to capture transient acoustic events while retaining sufficient frequency resolution.
    \item{Hop Ratio:} A 20\,\% overlap is maintained between consecutive frames (\textit{i.e.,} a hop ratio of 0.2). This partial overlap preserves temporal continuity and guards against losing information at frame boundaries—particularly important for detecting short, impulsive anomalies.
\end{itemize}
By transforming each spectrogram into a series of frames, we produce localized snapshots of the acoustic signal. These frames serve as input units for both training and evaluation, making the subsequent anomaly detection task more fine‐grained and sensitive to local transient events.

\begin{minipage}{0.95\linewidth}
\begin{lstlisting}[caption={Code to Generate Overlapping Frames from a Mel-Spectrogram},language=Python, label={lst:frame_generation}]
def generate_frames(mel_spectrogram, frame_size, hop_size):

    num_frames = (mel_spectrogram.shape[1] - frame_size) // hop_size + 1
    frames = np.zeros((num_frames, mel_spectrogram.shape[0], frame_size))
    for i in range(num_frames):
        start = i * hop_size
        frames[i] = mel_spectrogram[:, start:start + frame_size]
    return frames
\end{lstlisting}
\end{minipage}

The Python code in Listing~\ref{lst:frame_generation} illustrates how the generation of frames is implemented in practice. Each set of frames is then saved in \texttt{.npy} format for subsequent training and evaluation. By separating the raw audio loading from the downstream tasks, we ensure a clear workflow in which all models operate on the same preprocessed feature set.

\subsection{Anomaly Detection Models}

We selected three machine learning models due to their distinct methodological strengths, computational efficiency, and suitability to handle the complexities inherent to acoustic anomaly detection in hydropower:

\begin{itemize}
    \item \textbf{K-Means clustering}: has a minimal computational overhead and intuitive clustering based approach \cite{Ahn2021}, making it suitable for real-time applications with limited computational resources. 
    \item \textbf{Long Short-Term Memory Autoencoder (LSTM AE)}: a deep learning model capable of capturing temporal dependencies in sequential acoustic signals, excelling at identifying deviations from normal behaviors, as can be seen in \cite{Erniyazov_Kim_Jaleel_Lim_2024} where accuracy rates of 99\% were achieved when detecting anomalies.
    \item \textbf{One-Class Support Vector Machine (OC-SVM)}: a model that learns a tight decision boundary around normal (non-anomalous) data, flagging deviations from the decision boundary as anomalies \cite{meire2019, muller2020acoustic}. 
\end{itemize}

In addition, we compute each method’s outlier score on the validation set (normal‐only) to pick a percentile-based \textbf{anomaly threshold}. We evaluated multiple thresholds between the 5th and 95th percentiles, and selected the ones that maximizes the F1-score \cite{Yang1999}. 

%% file: sections/4_results.tex
\section{Experimental Evaluation} 
\label{sec:results}

This Section presents results of the proposed acoustic anomaly detection framework, evaluated on audio data collected with synthetic, induced anomalies, and in real operation mode. All audio was captured with a \textit{Samson Meteor MIC} at a 16 kHz sampling rate and stored in \texttt{.wav} format for consistency. We examine three ML models representing key industrial trade-offs: fast but simple methods, balanced approaches, and high-performance but resource-heavy solutions. 

\begin{figure}[h]
    \centering
    \includegraphics[width=0.78\columnwidth]{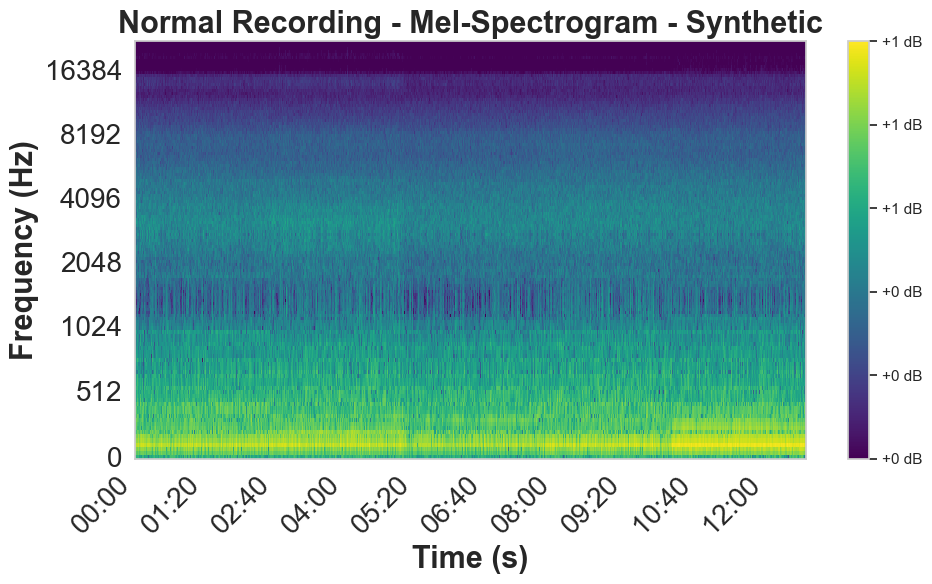}
    \vspace{-0.2cm}
    \includegraphics[width=0.78\columnwidth]{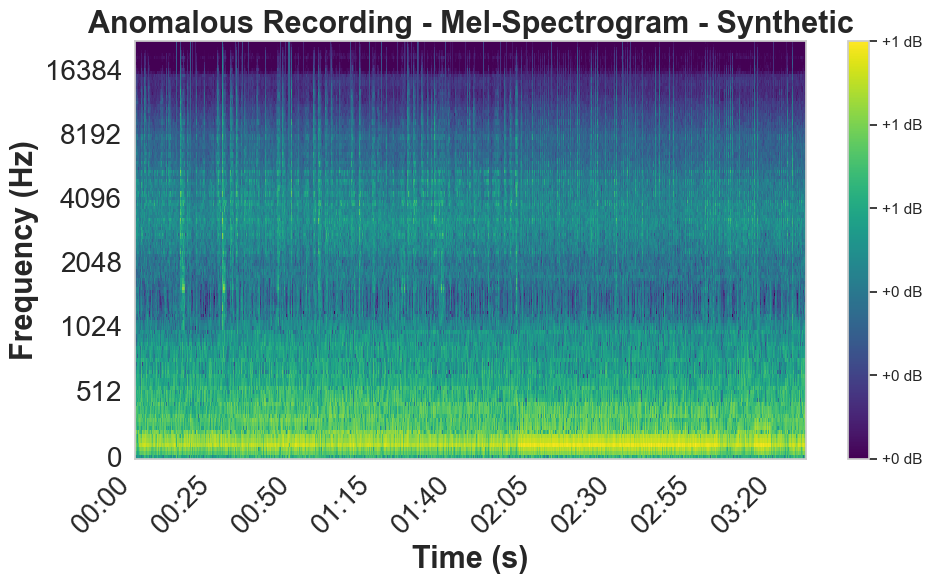}
    \caption{Mel-spectrograms for synthetic dataset: normal (top) vs. anomalous (bottom).}
    \label{fig:melspecSynthetic}
\end{figure}

\subsection{Induced Synthetic Anomalies}

The first dataset was recorded in an operational industrial machine and comprises:

\begin{figure*}[ht]
    \centering
    \includegraphics[width=0.25\textwidth]{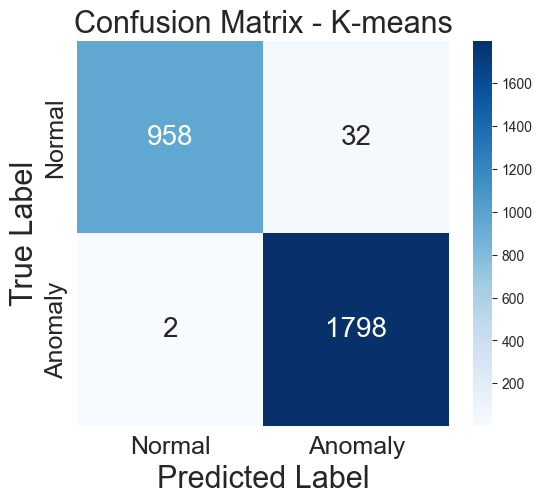}
    \hspace{0.02\textwidth}
    \includegraphics[width=0.25\textwidth]{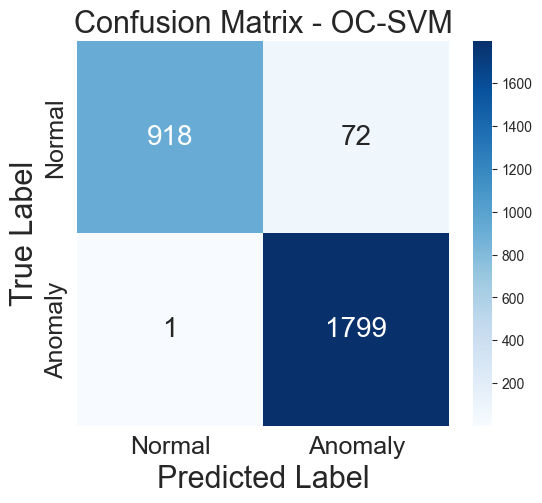}
    \hspace{0.02\textwidth}
    \includegraphics[width=0.25\textwidth]{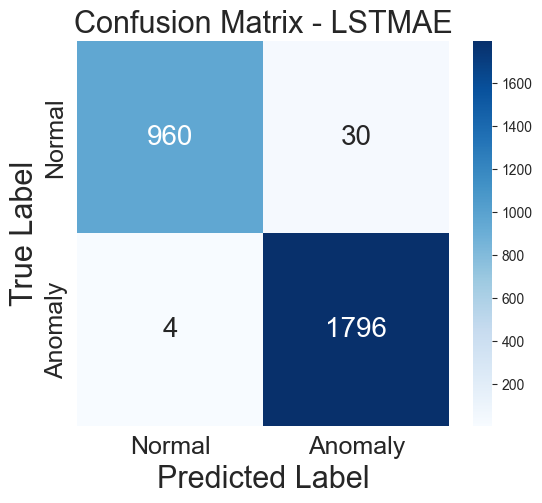}
    \caption{Confusion matrices: Induced (Synthetic) Anomalies Dataset}
    \label{fig:train_synthetic_confusion}
\end{figure*}

\begin{itemize} 
    \item \textbf{Normal operation (12 minutes 46 seconds)}: Equipment functioning under routine conditions, reflecting regular machine activity. 
    \item \textbf{Induced events (3 minutes 29 seconds)}: Artificial mechanical faults introduced via controlled impacts from a hammer striking a shovel, selected to resemble real‐world structural issues yet remain safe and repeatable. 
\end{itemize} 
These synthetic knocks allow replication of realistic fault phenomena and thus form the basis for testing the framework under near–real‐world conditions while preserving consistency in labeling.

The \textbf{Mel-spectrograms} of the synthetic anomaly and normal datasets are presented in Figure \ref{fig:melspecSynthetic}. The normal recording (top) shows smooth, constant energy distribution below 2000 HZ. The anomalous Mel-spectrogram (bottom) reveals multiple energy bursts with vertical stripes representing induced anomalies. It shows short high-energy bands in a high frequency range of above 2000 HZ, which correspond to the knocking sounds caused by a hammer hitting a shovel. Therefore, the normal operation noise is clearly visually distinguishable from the anomalous sound.

\begin{figure}[h]
    \centering
    \includegraphics[width=0.78\columnwidth]{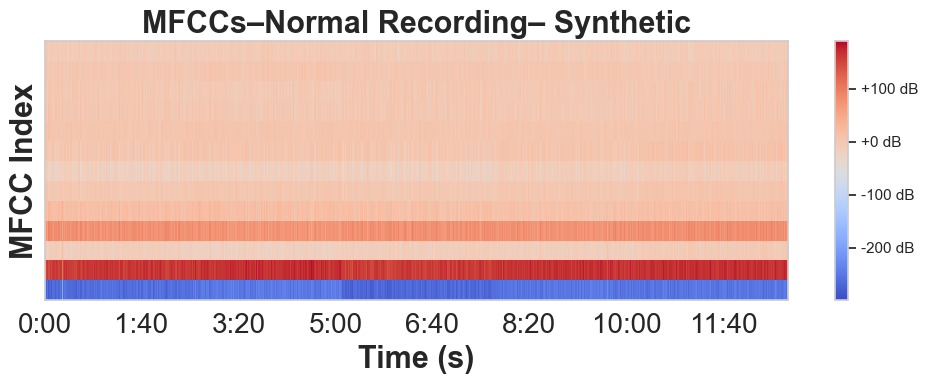}
    \vspace{-0.2cm}
    \includegraphics[width=0.78\columnwidth]{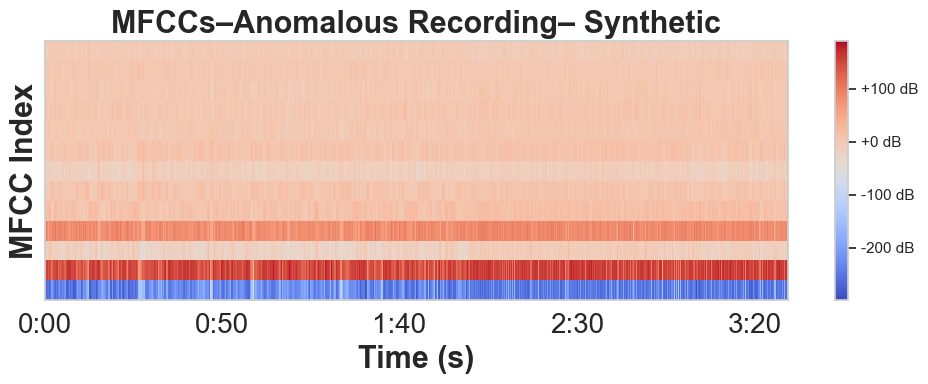}
    \caption{Comparison of Anomalous and Normal Recordings MFCCs}
    \label{fig:synthetic_mfccs_combined}
\end{figure}

Figure \ref{fig:synthetic_mfccs_combined} shows the \textbf{MFCC} coefficients for both recordings. The the normal recording (top) exhibits relatively uniform spectral patterns suggesting a stable envelope, while the anomalous recording (bottom) reveals sudden blanks in coefficients over short intervals, especially in coefficient 0 representing average log-energy. However, spectral patterns remain fairly similar between recordings, likely due to hydroelectric plant background noise.

\begin{figure}[h]
    \centering
    \includegraphics[width=0.78\columnwidth]{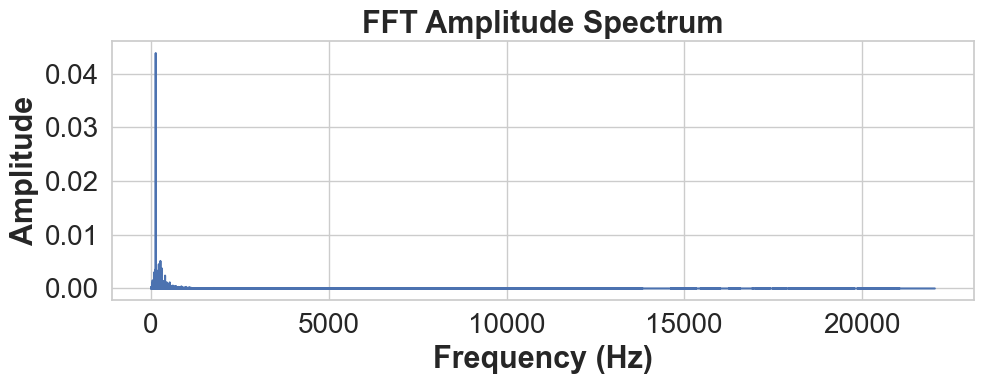}
    \vspace{-0.2cm}
    \includegraphics[width=0.78\columnwidth]{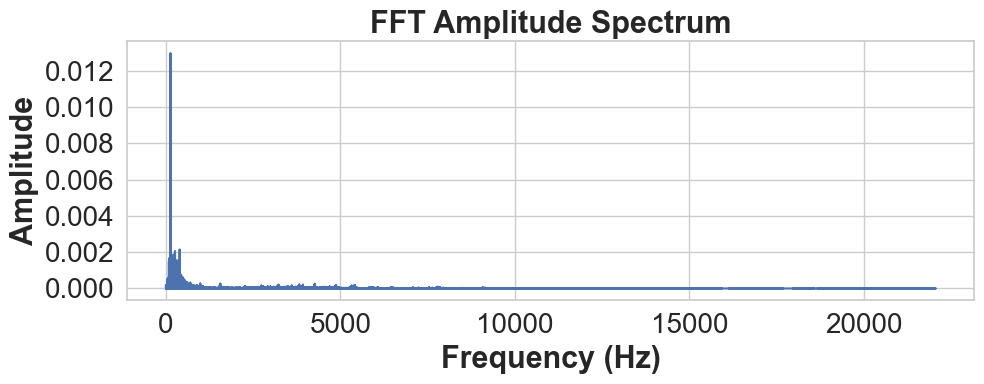}
    \caption{Comparison of Anomalous and Normal Recordings FFT Amplitudes}
    \label{fig:synthetic_fft_combined}
\end{figure}

Figure \ref{fig:synthetic_fft_combined} displays the \textbf{FFT Amplitude spectrum}, revealing amplitude distribution across frequencies. The normal recording (top) shows a concentrated spectral profile in the lower frequencies, while the anomalous recording (bottom) exhibits wider energy distribution with distinct amplitude spikes in higher frequencies, corresponding to induced anomalies. 

\noindent \textbf{Evaluation metrics}. Table~\ref{tab:synthetic_results} presents the \textbf{evaluation metrics} for the ML models on the synthetic anomaly dataset with focus on the K-Means, OC-SVM, and LSTM-AE. All models achieved near-perfect results, as reflected in the ROC AUC, precision, recall, and F1-scores. The \textbf{confusion matrices} in Figure~\ref{fig:train_synthetic_confusion} further highlight the strong classification performance. Overall, the LSTM-AE provides the most balanced outcome, maintaining both a low false positive rate (3.0\%) and high anomaly recall (99.8\%), making it well suited for deployment where precision and trust in anomaly alarms are essential. However, K-Means offers the fastest training and inference times, with performance metrics nearly matching those of LSTM-AE, and achieves the lowest overall misclassification count.

\begin{table}[htbp]
    \centering
    \caption{Evaluation metrics for the synthetic anomaly dataset.}
    \label{tab:synthetic_results}
    \resizebox{\columnwidth}{!}{%
        \begin{tabular}{|c|c|c|c|c|c|c|}
            \hline
            Method & Train Time (s) & ROC AUC & Precision & Recall & F1-Score & Inference Time (s) \\ \hline
            K-Means & 0.3700 & 0.9974 & 0.9825 & 0.9989 & 0.9906 & 0.0275 \\ \hline
            OC-SVM & 2.8284 & 0.9977 & 0.9615 & 0.9994 & 0.9801 & 2.1469 \\ \hline
            LSTM-AE & 32.8777 & 0.9995 & 0.9814 & 0.9972 & 0.9893 & 0.4765 \\ \hline
        \end{tabular}
     }
\end{table}

\subsection{Real-world Operation}

The second dataset was recorded in an operational settings and comprises: 

\begin{itemize} 
    \item \textbf{Normal operation (59 minutes 53 seconds)}: equipment functioning under routine conditions, reflecting regular machine activity. 
    \item \textbf{Transient acoustic event (5 seconds within 59 minutes)}: detected during the transition phase, \textit{i.e.,} electricity generation to water pump. 
\end{itemize} 

\begin{figure}[b]
    \centering
    \includegraphics[width=0.78\columnwidth]{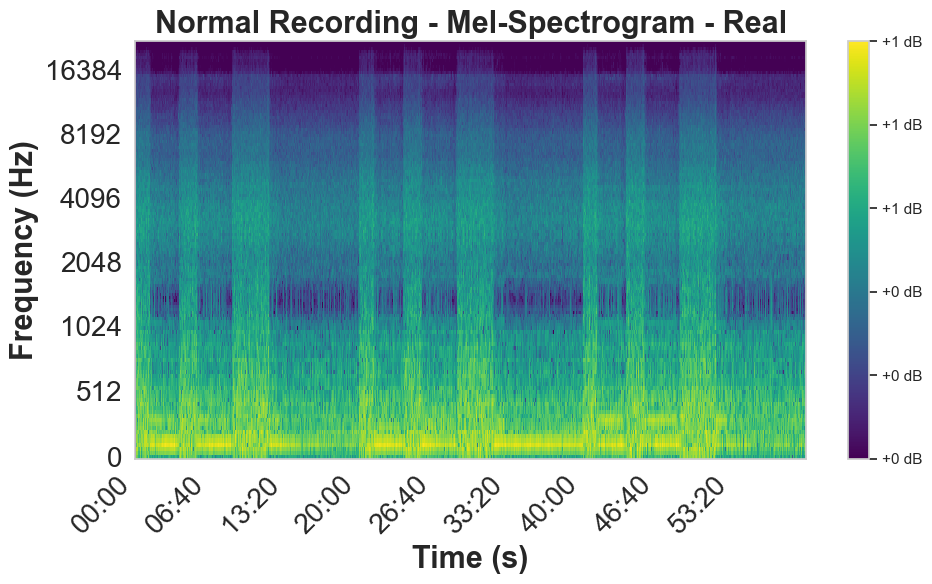}
    \vspace{-0.2cm}
    \includegraphics[width=0.78\columnwidth]{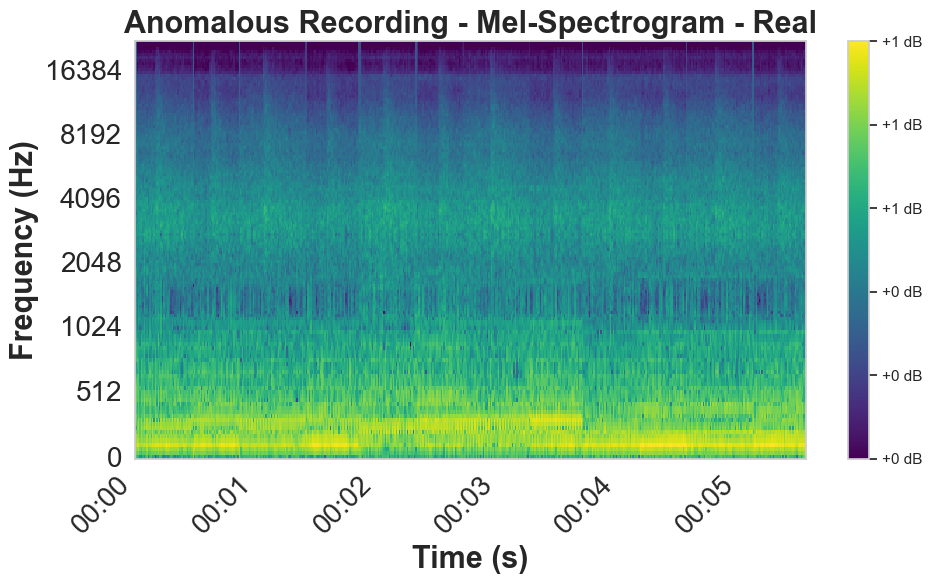}
    \caption{Mel-spectrograms for real dataset: normal (top) vs. anomalous (bottom).}
    \label{fig:melspecReal}
\end{figure}

\begin{figure*}[ht]
    \centering
    \includegraphics[width=0.25\textwidth]{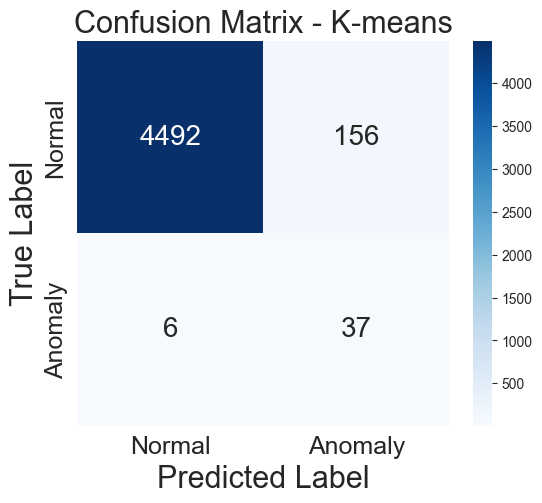}
    \hspace{0.02\textwidth}
    \includegraphics[width=0.25\textwidth]{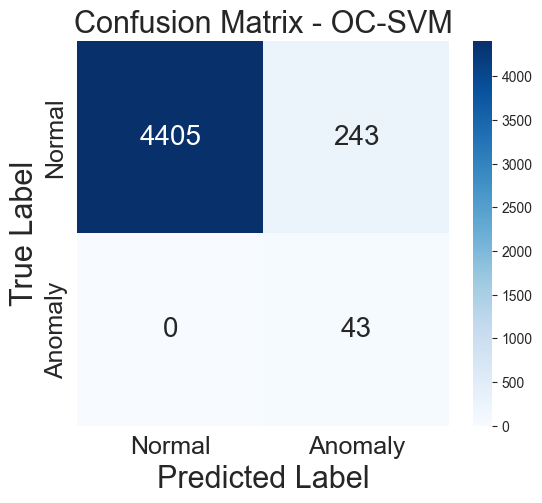}
    \hspace{0.02\textwidth}
    \includegraphics[width=0.25\textwidth]{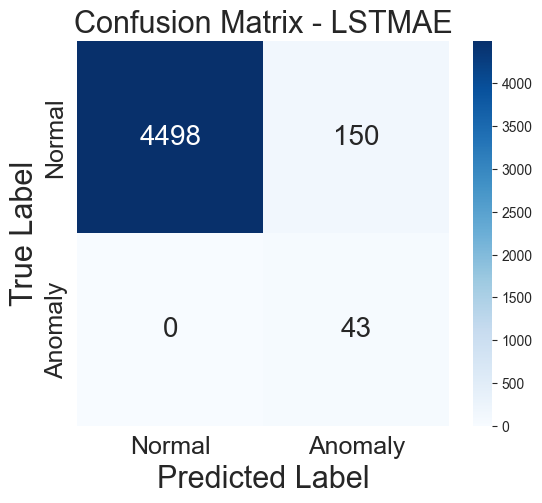}
    \caption{Confusion matrices: Real Industrial Dataset}
    \label{fig:train_real_confusion}
\end{figure*}

According to operational insights from plant personnel, acoustic anomalies typically occur during specific operational transitions: startup sequences, mode switching between pumping and generation phases, and load adjustment periods. These transient events are \textbf{not} indicative of equipment failure but rather reflect the dynamic behavior of turbine systems designed to operate in dual modes. Also, they are brief by nature, as prolonged anomalies would indicate actual equipment malfunction or damage rather than normal operational variations. The short duration of real anomalies (5 seconds within 59 minutes recording) and the constrained dataset size reflect this operational reality where brief acoustic deviations are characteristic of healthy industrial systems, combined with the practical challenges of extended data collection in critical infrastructure environments.

The \textbf{Mel-spectrogram} of the real anomaly dataset as well as the normal and anomalous frames are shown in Figure~\ref{fig:melspecReal}. The operating modes are clearly visible through pronounced energy shifts, but these do not necessarily correspond to anomalies since normal mode transitions are prolonged with uniform frequency distribution. Conversely, the anomalous recording (bottom) exhibits inconsistent frequency changes without apparent pattern throughout the recording.

\begin{figure}[h]
    \centering
    \includegraphics[width=0.78\columnwidth]{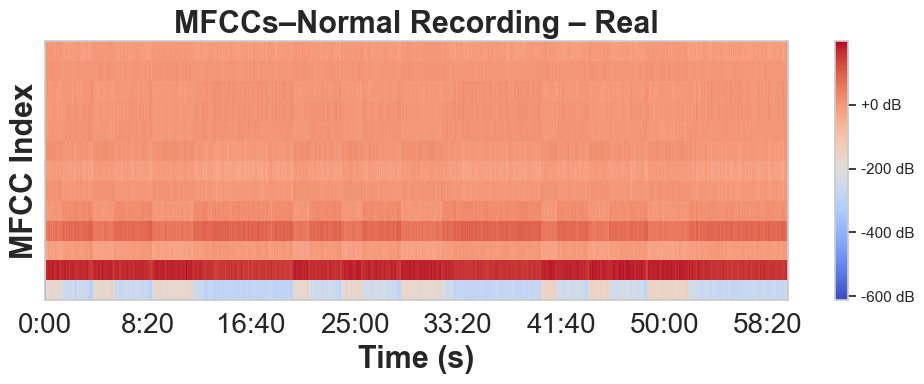}
    \vspace{-0.2cm}
    \includegraphics[width=0.78\columnwidth]{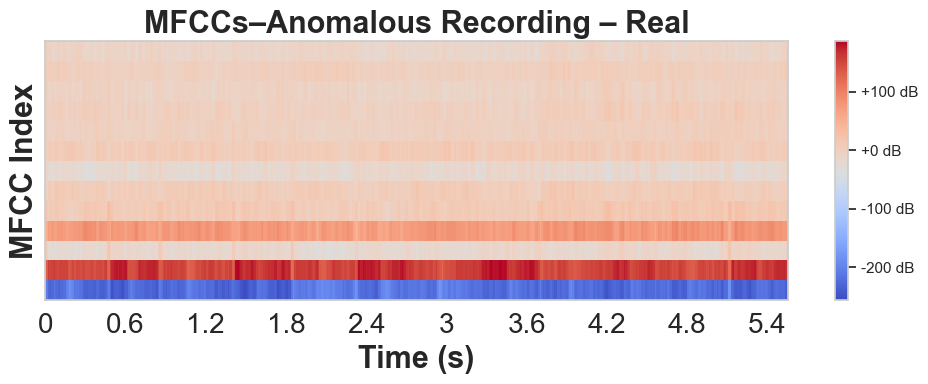}
    \caption{Comparison of Real Recordings MFCCs}
    \label{fig:real_mfccs_combined}
\end{figure}

The previous observations can also be seen in Figure \ref{fig:real_mfccs_combined}, which shows the \textbf{MFCCs} of both recordings. The normal MFCC (top) shows consistent color bands in most coefficients in the normal recording indicating a stable spectral envelope. Apart from the evident the changes in operating modes in coefficient 0, each coefficient contains a stable spectral envelope. In contrast, the anomalous recording (bottom) exhibits higher variability in the MFCC coefficients. The coefficients 1-12 fluctuate more intensely during anomalies, showing unstable spectral patterns. Additionally, coefficient 0 shows a lower log-energy suggesting reduced energy in abnormal sounds.

\begin{figure}[h]
    \centering
    \includegraphics[width=0.78\columnwidth]{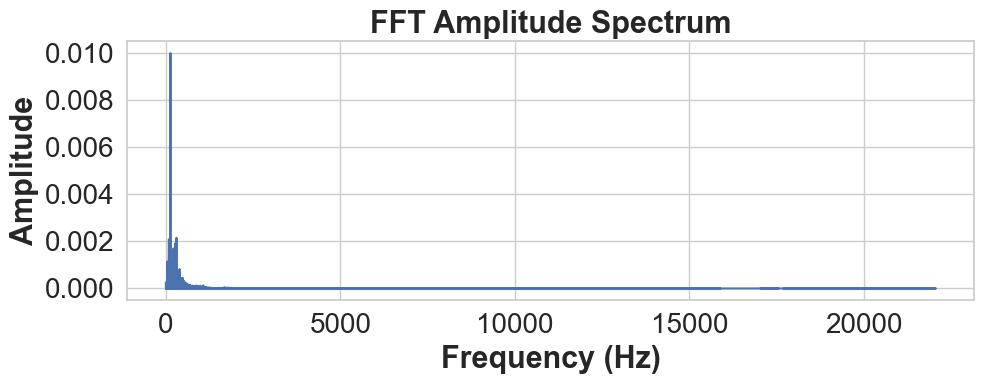}
    \vspace{-0.2cm}
    \includegraphics[width=0.78\columnwidth]{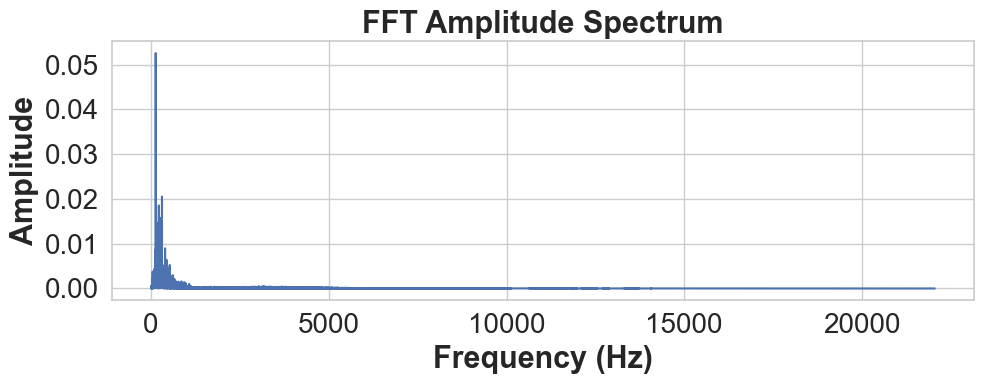}
    \caption{Comparison of Real Recordings FFT Amplitude}
    \label{fig:real_fft_amplitudes_combined}
\end{figure}

Lastly, Figure \ref{fig:real_fft_amplitudes_combined} shows a comparison of the FFT Amplitude spectrum of both recordings. The amplitude and frequency distribution of the anomalous FFT (bottom) is significantly broader compared to the normal recording (top). The frequency distribution for the normal is much more compact, representing normal operating conditions. Furthermore, the amplitude peak in the normal recording is 0.010dB compared to a higher peak in the anomalous recording of over 0.050dB. This is due to the longer audio file for the normal recording and more distributed acoustic signals in contrast to the short and noisy signals in the anomalous recording. 

\begin{table}[htbp]
    \centering
    \caption{Evaluation metrics for the real world dataset.}
    \label{tab:real_results}
    \resizebox{\columnwidth}{!}{%
        \begin{tabular}{|c|c|c|c|c|c|c|}
            \hline
            Method & Train Time (s) & ROC AUC & Precision & Recall & F1-Score & Inference Time (s) \\ \hline
            K-Means & 5.25 & 0.962 & 0.192 & 0.861 & 0.314 & 0.043 \\ \hline
            OC-SVM & 55.54 & 0.998 & 0.150 & 1.000 & 0.261 & 12.17 \\ \hline
            LSTM-AE & 145.25 & 0.997 & 0.219 & 1.000 & 0.360 & 0.867 \\ \hline
        \end{tabular}
    }
\end{table}

\noindent \textbf{Evaluation metrics}. In the real-world scenario, the confusion matrices (cf. Figure \ref{fig:train_real_confusion} confirm that all models successfully detected the transient event during the transition phase, with LSTM-AE and OC-SVM achieving perfect recall. LSTM-AE correctly identified all 43 transient events samples with 150 false positives, achieving the highest F1-score (0.360). OC-SVM also detected all anomalies but misclassified 243 normal samples (F1-score: 0.261). K-Means missed 6 anomalies with 156 false positives, achieving the lowest F1-score (0.314). Per Table~\ref{tab:real_results}, K-Means remains fastest for training and inference, while OC-SVM is significantly slower at inference.

\subsection{Comparative Analysis}


The experimental evaluation outlines the crucial role of effective audio feature extraction. Mel-spectograms and MFCCs provided key inputs by distinguishing subtle frequency variations, aligning with our exploratory data analysis. This preprocessing stage directly influenced model performance in detecting anomalies despite typical hydropower challenges like background noise and signal reflections.

Model selection involves notable \textbf{trade-offs} among accuracy, computational efficiency, and explainability:
\begin{itemize}
    \item \textbf{K-means} offered fast inference (0.027–0.043s) for real-time applications but limited robustness to overlapping acoustic patterns (precision: 0.192).
    \item \textbf{OC-SVM} demonstrated strong detection (ROC AUC: 0.9976 – 0.998) with moderate computational overhead (training time: 2.8–55.5s) and intuitive boundary-based interpretability.
    \item \textbf{LSTM Autoencoder (AE)} captured complex temporal dependencies effectively (ROC AUC: 0.997–0.999) but required significantly higher computational complexity (training time: 145s). Its performance justifies deployment where accuracy is paramount, though explainability and maintenance may pose practical challenges.
\end{itemize}

%% file: sections/6_conclusions.tex
\section{Conclusions and Future Work} \label{sec:conclusion}






This paper presented an acoustic anomaly detection pipeline for predictive maintenance in PSHP power plants. We benchmark three ML models (K-Means, OC-SVM, and LSTM-AE) on real-world acoustic datasets from Rodundwerk II, analyzing audio feature selection and model trade-offs between accuracy, computational cost, and interpretability. OC-SVM achieved optimal efficiency-accuracy balance, while LSTM-AE excelled at detecting subtle temporal anomalies despite higher computational demands.

Future work includes developing a multi-modal anomaly detection framework that integrates acoustic signals with vibration data and operational parameters (\textit{e.g.,} rotational speed, bearing temperature, power output). Additionally, collecting expanded datasets from multiple microphone arrays positioned across different floors of the plant will enable extending detection capabilities from 2D to 3D, incorporating localization of anomalies.

%% file: sample-sigconf.bbl

\begin{thebibliography}{32}


\ifx \showCODEN    \undefined \def \showCODEN     #1{\unskip}     \fi
\ifx \showISBNx    \undefined \def \showISBNx     #1{\unskip}     \fi
\ifx \showISBNxiii \undefined \def \showISBNxiii  #1{\unskip}     \fi
\ifx \showISSN     \undefined \def \showISSN      #1{\unskip}     \fi
\ifx \showLCCN     \undefined \def \showLCCN      #1{\unskip}     \fi
\ifx \shownote     \undefined \def \shownote      #1{#1}          \fi
\ifx \showarticletitle \undefined \def \showarticletitle #1{#1}   \fi
\ifx \showURL      \undefined \def \showURL       {\relax}        \fi
\providecommand\bibfield[2]{#2}
\providecommand\bibinfo[2]{#2}
\providecommand\natexlab[1]{#1}
\providecommand\showeprint[2][]{arXiv:#2}

\bibitem[Ahn and Yeo(2021)]%
        {Ahn2021}
\bibfield{author}{\bibinfo{person}{Hyojung Ahn} {and} \bibinfo{person}{Inchoon Yeo}.} \bibinfo{year}{2021}\natexlab{}.
\newblock \showarticletitle{Deep-Learning-Based Approach to Anomaly Detection Techniques for Large Acoustic Data in Machine Operation}.
\newblock \bibinfo{journal}{\emph{Sensors}} \bibinfo{volume}{21}, \bibinfo{number}{5446} (\bibinfo{year}{2021}).
\newblock
\href{https://doi.org/10.3390/s21165446}{doi:\nolinkurl{10.3390/s21165446}}


\bibitem[Bayram et~al\mbox{.}(2021)]%
        {Bayram2021}
\bibfield{author}{\bibinfo{person}{Barış Bayram}, \bibinfo{person}{Taha~Berkay Duman}, {and} \bibinfo{person}{Gökhan Ince}.} \bibinfo{year}{2021}\natexlab{}.
\newblock \showarticletitle{Real time detection of acoustic anomalies in industrial processes using sequential autoencoders}.
\newblock \bibinfo{journal}{\emph{Expert Systems}} \bibinfo{volume}{38}, \bibinfo{number}{1} (\bibinfo{year}{2021}), \bibinfo{pages}{e12564}.
\newblock
\showISSN{0266-4720}
\href{https://doi.org/10.1111/exsy.12564}{doi:\nolinkurl{10.1111/exsy.12564}}


\bibitem[Breunig et~al\mbox{.}(2000)]%
        {Breunig2000}
\bibfield{author}{\bibinfo{person}{M.~M. Breunig}, \bibinfo{person}{H.~P. Kriegel}, \bibinfo{person}{R.~T. Ng}, {and} \bibinfo{person}{J. Sander}.} \bibinfo{year}{2000}\natexlab{}.
\newblock \showarticletitle{LOF: Identifying density-based local outliers}.
\newblock \bibinfo{journal}{\emph{ACM SIGMOD Record}} \bibinfo{volume}{29}, \bibinfo{number}{2} (\bibinfo{year}{2000}), \bibinfo{pages}{93--104}.
\newblock
\href{https://doi.org/10.1145/335191.335388}{doi:\nolinkurl{10.1145/335191.335388}}


\bibitem[Coelho et~al\mbox{.}(2022)]%
        {Coelho2022DeepAutoencoders}
\bibfield{author}{\bibinfo{person}{Gabriel Coelho}, \bibinfo{person}{Lu\'{i}s~Miguel Matos}, \bibinfo{person}{Pedro~Jos\'{e} Pereira}, \bibinfo{person}{Andr\'{e} Ferreira}, \bibinfo{person}{Andr\'{e} Pilastri}, {and} \bibinfo{person}{Paulo Cortez}.} \bibinfo{year}{2022}\natexlab{}.
\newblock \showarticletitle{Deep Autoencoders for Acoustic Anomaly Detection: Experiments with Working Machine and In-Vehicle Audio}.
\newblock \bibinfo{journal}{\emph{Reposit\'{o}riUM}} (\bibinfo{year}{2022}).
\newblock
\urldef\tempurl%
\url{https://repositorium.sdum.uminho.pt/bitstream/1822/81433/1/aeaad2.pdf}
\showURL{%
\tempurl}
\newblock
\shownote{Accessed: 2025-02-16}.


\bibitem[Cohen(1995)]%
        {cohen1995time}
\bibfield{author}{\bibinfo{person}{Leon Cohen}.} \bibinfo{year}{1995}\natexlab{}.
\newblock \bibinfo{booktitle}{\emph{Time-frequency analysis}}. Vol.~\bibinfo{volume}{778}.
\newblock \bibinfo{publisher}{Prentice Hall PTR New Jersey}.
\newblock


\bibitem[Davis and Mermelstein(1980)]%
        {1163420}
\bibfield{author}{\bibinfo{person}{S. Davis} {and} \bibinfo{person}{P. Mermelstein}.} \bibinfo{year}{1980}\natexlab{}.
\newblock \showarticletitle{Comparison of parametric representations for monosyllabic word recognition in continuously spoken sentences}.
\newblock \bibinfo{journal}{\emph{IEEE Transactions on Acoustics, Speech, and Signal Processing}} \bibinfo{volume}{28}, \bibinfo{number}{4} (\bibinfo{year}{1980}), \bibinfo{pages}{357--366}.
\newblock
\href{https://doi.org/10.1109/TASSP.1980.1163420}{doi:\nolinkurl{10.1109/TASSP.1980.1163420}}


\bibitem[Di~Fiore et~al\mbox{.}(2022)]%
        {diFiore2022}
\bibfield{author}{\bibinfo{person}{E. Di~Fiore}, \bibinfo{person}{A. Ferraro}, \bibinfo{person}{A. Galli}, \bibinfo{person}{V. Moscato}, {and} \bibinfo{person}{G. Sperlì}.} \bibinfo{year}{2022}\natexlab{}.
\newblock \showarticletitle{An anomalous sound detection methodology for predictive maintenance}.
\newblock \bibinfo{journal}{\emph{Expert Systems with Applications}}  \bibinfo{volume}{209} (\bibinfo{year}{2022}), \bibinfo{pages}{118324}.
\newblock
\href{https://doi.org/10.1016/j.eswa.2022.118324}{doi:\nolinkurl{10.1016/j.eswa.2022.118324}}


\bibitem[Duman et~al\mbox{.}(2020)]%
        {duman2020}
\bibfield{author}{\bibinfo{person}{T.~B. Duman}, \bibinfo{person}{B. Bayram}, {and} \bibinfo{person}{G. İnce}.} \bibinfo{year}{2020}\natexlab{}.
\newblock \showarticletitle{Acoustic Anomaly Detection Using Convolutional Autoencoders in Industrial Processes}.
\newblock In \bibinfo{booktitle}{\emph{14th International Conference on Soft Computing Models in Industrial and Environmental Applications (SOCO 2019)}}, \bibfield{editor}{\bibinfo{person}{F.~Martínez~Álvarez}, \bibinfo{person}{A.~Troncoso~Lora}, \bibinfo{person}{J.~A. Sáez~Muñoz}, \bibinfo{person}{H.~Quintián}, {and} \bibinfo{person}{E.~Corchado}} (Eds.). \bibinfo{publisher}{Springer International Publishing}, \bibinfo{pages}{432--442}.
\newblock
\href{https://doi.org/10.1007/978-3-030-20055-8_41}{doi:\nolinkurl{10.1007/978-3-030-20055-8_41}}


\bibitem[E-control(2025)]%
        {EControl.}
\bibfield{author}{\bibinfo{person}{E-control}.} \bibinfo{year}{2025}\natexlab{}.
\newblock \bibinfo{title}{Aktueller Marktpreis gem{\"a}{\ss} {\S} 41 Abs. 1 {\"O}kostromgesetz 2012}.
\newblock
\urldef\tempurl%
\url{https://www.e-control.at/industrie/oeko-energie/oekostrommarkt/marktpreise-gem-paragraph-20#:~:text=Dieser%20Marktpreis%20betr%C3%A4gt%20demzufolge%2097,Quartal%202024).}
\showURL{%
\tempurl}


\bibitem[Erniyazov et~al\mbox{.}(2024)]%
        {Erniyazov_Kim_Jaleel_Lim_2024}
\bibfield{author}{\bibinfo{person}{Sarvarbek Erniyazov}, \bibinfo{person}{Yongmin Kim}, \bibinfo{person}{M. Jaleel}, {and} \bibinfo{person}{Chang~Gyoon Lim}.} \bibinfo{year}{2024}\natexlab{}.
\newblock \showarticletitle{Comprehensive Analysis and Improved Techniques for Anomaly Detection in Time Series Data with Autoencoder Models}.
\newblock \bibinfo{journal}{\emph{International Journal on Advanced Science, Engineering and Information Technology}} \bibinfo{volume}{14}, \bibinfo{number}{6} (\bibinfo{date}{Dec.} \bibinfo{year}{2024}), \bibinfo{pages}{1861–1867}.
\newblock
\href{https://doi.org/10.18517/ijaseit.14.6.20451}{doi:\nolinkurl{10.18517/ijaseit.14.6.20451}}


\bibitem[Ferraro et~al\mbox{.}(2023)]%
        {Ferraro2023UnsupervisedAnomaly}
\bibfield{author}{\bibinfo{person}{A. Ferraro}, \bibinfo{person}{A. Galli}, \bibinfo{person}{V.L. Gatta}, \bibinfo{person}{V. Moscato}, \bibinfo{person}{M. Postiglione}, \bibinfo{person}{G. Sperl{\`i}}, {and} \bibinfo{person}{F. Moscato}.} \bibinfo{year}{2023}\natexlab{}.
\newblock \showarticletitle{Unsupervised Anomaly Detection in Predictive Maintenance using Sound Data}.
\newblock \bibinfo{journal}{\emph{CEUR Workshop Proceedings}}  \bibinfo{volume}{3478} (\bibinfo{year}{2023}), \bibinfo{pages}{449--458}.
\newblock


\bibitem[illwerke vkw.(2025)]%
        {illwerkeRodundwerk}
\bibfield{author}{\bibinfo{person}{illwerke vkw.}} \bibinfo{year}{2025}\natexlab{}.
\newblock \bibinfo{title}{{Rodundwerk II}}.
\newblock
\urldef\tempurl%
\url{https://www.illwerkevkw.at/rodundwerk-ii}
\showURL{%
\tempurl}


\bibitem[Khanjari et~al\mbox{.}(2024)]%
        {Khanjari2024}
\bibfield{author}{\bibinfo{person}{M. Khanjari}, \bibinfo{person}{A. Azarfar}, \bibinfo{person}{M.~H. Abardeh}, {and} \bibinfo{person}{E. Alibeiki}.} \bibinfo{year}{2024}\natexlab{}.
\newblock \showarticletitle{Anomalous sound detection for machine condition monitoring using 3D tensor representation of sound and 3D deep convolutional neural network}.
\newblock \bibinfo{journal}{\emph{Multimedia Tools and Applications}} \bibinfo{volume}{83}, \bibinfo{number}{15} (\bibinfo{year}{2024}), \bibinfo{pages}{44101--44119}.
\newblock
\href{https://doi.org/10.1007/s11042-023-17043-9}{doi:\nolinkurl{10.1007/s11042-023-17043-9}}


\bibitem[Kishor et~al\mbox{.}(2007)]%
        {kishor2007review}
\bibfield{author}{\bibinfo{person}{Nand Kishor}, \bibinfo{person}{RP Saini}, {and} \bibinfo{person}{SP Singh}.} \bibinfo{year}{2007}\natexlab{}.
\newblock \showarticletitle{A review on hydropower plant models and control}.
\newblock \bibinfo{journal}{\emph{Renewable and Sustainable Energy Reviews}} \bibinfo{volume}{11}, \bibinfo{number}{5} (\bibinfo{year}{2007}), \bibinfo{pages}{776--796}.
\newblock


\bibitem[König et~al\mbox{.}(2021)]%
        {konig2021}
\bibfield{author}{\bibinfo{person}{F. König}, \bibinfo{person}{C. Sous}, \bibinfo{person}{A. Ouald~Chaib}, {and} \bibinfo{person}{G. Jacobs}.} \bibinfo{year}{2021}\natexlab{}.
\newblock \showarticletitle{Machine learning-based anomaly detection and classification of acoustic emission events for wear monitoring in sliding bearing systems}.
\newblock \bibinfo{journal}{\emph{Tribology International}}  \bibinfo{volume}{155} (\bibinfo{year}{2021}), \bibinfo{pages}{106811}.
\newblock
\href{https://doi.org/10.1016/j.triboint.2020.106811}{doi:\nolinkurl{10.1016/j.triboint.2020.106811}}


\bibitem[Linnhoff-Popien et~al\mbox{.}(2021)]%
        {muller2020acoustic}
\bibfield{author}{\bibinfo{person}{Claudia Linnhoff-Popien}, \bibinfo{person}{Steffen Illium}, \bibinfo{person}{Fabian Ritz}, {and} \bibinfo{person}{Robert M\"uller}.} \bibinfo{year}{2021}\natexlab{}.
\newblock \showarticletitle{Acoustic Anomaly Detection for Machine Sounds based on Image Transfer Learning}.
\newblock In \bibinfo{booktitle}{\emph{Proceedings of the 13th International Conference on Agents and Artificial Intelligence (Volume 2)}}, \bibfield{editor}{\bibinfo{person}{Ana~Paula Rocha}, \bibinfo{person}{Luc Steels}, {and} \bibinfo{person}{Jaap van~den Herik}} (Eds.). \bibinfo{publisher}{SciTePress}, \bibinfo{address}{Set\'ubal, Portuagl}, \bibinfo{pages}{49--56}.
\newblock


\bibitem[Lo~Scudo et~al\mbox{.}(2023)]%
        {loScudo2023}
\bibfield{author}{\bibinfo{person}{F. Lo~Scudo}, \bibinfo{person}{E. Ritacco}, \bibinfo{person}{L. Caroprese}, {and} \bibinfo{person}{G. Manco}.} \bibinfo{year}{2023}\natexlab{}.
\newblock \showarticletitle{Audio-based anomaly detection on edge devices via self-supervision and spectral analysis}.
\newblock \bibinfo{journal}{\emph{Journal of Intelligent Information Systems}} \bibinfo{volume}{61}, \bibinfo{number}{3} (\bibinfo{year}{2023}), \bibinfo{pages}{765--793}.
\newblock
\href{https://doi.org/10.1007/s10844-023-00792-2}{doi:\nolinkurl{10.1007/s10844-023-00792-2}}


\bibitem[Malcolm(1998)]%
        {malcolm1998auditory}
\bibfield{author}{\bibinfo{person}{Slaney Malcolm}.} \bibinfo{year}{1998}\natexlab{}.
\newblock \showarticletitle{Auditory toolbox version 2}.
\newblock \bibinfo{journal}{\emph{https://engineering. purdue. edu/\~{} malcolm/interval/1998-010/}} (\bibinfo{year}{1998}).
\newblock


\bibitem[McFee et~al\mbox{.}(2015)]%
        {mcfee2015librosa}
\bibfield{author}{\bibinfo{person}{Brian McFee}, \bibinfo{person}{Colin Raffel}, \bibinfo{person}{Dawen Liang}, \bibinfo{person}{Daniel~PW Ellis}, \bibinfo{person}{Matt McVicar}, \bibinfo{person}{Eric Battenberg}, {and} \bibinfo{person}{Oriol Nieto}.} \bibinfo{year}{2015}\natexlab{}.
\newblock \showarticletitle{librosa: Audio and music signal analysis in python.}. In \bibinfo{booktitle}{\emph{SciPy}}. \bibinfo{pages}{18--24}.
\newblock


\bibitem[Meire and Karsmakers(2019)]%
        {meire2019}
\bibfield{author}{\bibinfo{person}{M. Meire} {and} \bibinfo{person}{P. Karsmakers}.} \bibinfo{year}{2019}\natexlab{}.
\newblock \showarticletitle{Comparison of Deep Autoencoder Architectures for Real-time Acoustic Based Anomaly Detection in Assets}. In \bibinfo{booktitle}{\emph{2019 10th IEEE International Conference on Intelligent Data Acquisition and Advanced Computing Systems (IDAACS)}}, Vol.~\bibinfo{volume}{2}. \bibinfo{pages}{786--790}.
\newblock
\href{https://doi.org/10.1109/IDAACS.2019.8924301}{doi:\nolinkurl{10.1109/IDAACS.2019.8924301}}


\bibitem[Mobley(2002)]%
        {Mobley2002}
\bibfield{author}{\bibinfo{person}{R.~Keith Mobley}.} \bibinfo{year}{2002}\natexlab{}.
\newblock \showarticletitle{1 - Impact of Maintenance}.
\newblock In \bibinfo{booktitle}{\emph{An Introduction to Predictive Maintenance (Second Edition)} (\bibinfo{edition}{second edition} ed.)}, \bibfield{editor}{\bibinfo{person}{R.~Keith Mobley}} (Ed.). \bibinfo{publisher}{Butterworth-Heinemann}, \bibinfo{address}{Burlington}, \bibinfo{pages}{1--22}.
\newblock
\showISBNx{978-0-7506-7531-4}
\href{https://doi.org/10.1016/B978-075067531-4/50001-4}{doi:\nolinkurl{10.1016/B978-075067531-4/50001-4}}


\bibitem[P{\'e}rez-D{\'\i}az et~al\mbox{.}(2015)]%
        {perez2015trends}
\bibfield{author}{\bibinfo{person}{Juan~I P{\'e}rez-D{\'\i}az}, \bibinfo{person}{Manuel Chazarra}, \bibinfo{person}{Javier Garc{\'\i}a-Gonz{\'a}lez}, \bibinfo{person}{Giovanna Cavazzini}, {and} \bibinfo{person}{Anna Stoppato}.} \bibinfo{year}{2015}\natexlab{}.
\newblock \showarticletitle{Trends and challenges in the operation of pumped-storage hydropower plants}.
\newblock \bibinfo{journal}{\emph{Renewable and Sustainable Energy Reviews}}  \bibinfo{volume}{44} (\bibinfo{year}{2015}), \bibinfo{pages}{767--784}.
\newblock


\bibitem[Portnoff(1980)]%
        {1163359}
\bibfield{author}{\bibinfo{person}{M. Portnoff}.} \bibinfo{year}{1980}\natexlab{}.
\newblock \showarticletitle{Time-frequency representation of digital signals and systems based on short-time Fourier analysis}.
\newblock \bibinfo{journal}{\emph{IEEE Transactions on Acoustics, Speech, and Signal Processing}} \bibinfo{volume}{28}, \bibinfo{number}{1} (\bibinfo{year}{1980}), \bibinfo{pages}{55--69}.
\newblock
\href{https://doi.org/10.1109/TASSP.1980.1163359}{doi:\nolinkurl{10.1109/TASSP.1980.1163359}}


\bibitem[Purohit et~al\mbox{.}(2019)]%
        {purohit2019mimii}
\bibfield{author}{\bibinfo{person}{Harsh Purohit}, \bibinfo{person}{Ryo Tanabe}, \bibinfo{person}{Takashi Ichige}, \bibinfo{person}{Tomoya Endo}, \bibinfo{person}{Yasunori Nikaido}, \bibinfo{person}{Kohei Suefusa}, {and} \bibinfo{person}{Yasue Kawaguchi}.} \bibinfo{year}{2019}\natexlab{}.
\newblock \showarticletitle{{MIMII} dataset: Sound dataset for malfunctioning industrial machine investigation and inspection}. In \bibinfo{booktitle}{\emph{Proceedings of the Detection and Classification of Acoustic Scenes and Events 2019 Workshop (DCASE2019)}}.
\newblock
\href{https://doi.org/10.33682/m76f-d618}{doi:\nolinkurl{10.33682/m76f-d618}}


\bibitem[Raju et~al\mbox{.}(2020)]%
        {9214160}
\bibfield{author}{\bibinfo{person}{V~N~Ganapathi Raju}, \bibinfo{person}{K~Prasanna Lakshmi}, \bibinfo{person}{Vinod~Mahesh Jain}, \bibinfo{person}{Archana Kalidindi}, {and} \bibinfo{person}{V Padma}.} \bibinfo{year}{2020}\natexlab{}.
\newblock \showarticletitle{Study the Influence of Normalization/Transformation process on the Accuracy of Supervised Classification}. In \bibinfo{booktitle}{\emph{2020 Third International Conference on Smart Systems and Inventive Technology (ICSSIT)}}. \bibinfo{pages}{729--735}.
\newblock
\href{https://doi.org/10.1109/ICSSIT48917.2020.9214160}{doi:\nolinkurl{10.1109/ICSSIT48917.2020.9214160}}


\bibitem[{ScienceDirect}(2025)]%
        {sciencedirect_audio_segmentation}
\bibfield{author}{\bibinfo{person}{{ScienceDirect}}.} \bibinfo{year}{2025}\natexlab{}.
\newblock \bibinfo{title}{Audio Segmentation}.
\newblock
\urldef\tempurl%
\url{https://www.sciencedirect.com/topics/engineering/audio-segmentation}
\showURL{%
\tempurl}
\newblock
\shownote{Accessed: 2025-02-13}.


\bibitem[Siemens(2023)]%
        {siemens2023downtime}
\bibfield{author}{\bibinfo{person}{Siemens}.} \bibinfo{year}{2023}\natexlab{}.
\newblock \bibinfo{title}{The True Cost of Downtime: Identify, Avoid, and Overcome}.
\newblock
\urldef\tempurl%
\url{https://assets.new.siemens.com/siemens/assets/api/uuid:3d606495-dbe0-43e4-80b1-d04e27ada920/dics-b10153-00-7600truecostofdowntime2022-144.pdf}
\showURL{%
\tempurl}
\newblock
\shownote{Digital Industries, Customer Services}.


\bibitem[Tagawa et~al\mbox{.}(2021)]%
        {tagawa2021}
\bibfield{author}{\bibinfo{person}{Y. Tagawa}, \bibinfo{person}{R. Maskeliūnas}, {and} \bibinfo{person}{R. Damaševičius}.} \bibinfo{year}{2021}\natexlab{}.
\newblock \showarticletitle{Acoustic anomaly detection of mechanical failures in noisy real-life factory environments}.
\newblock \bibinfo{journal}{\emph{Electronics}} \bibinfo{volume}{10}, \bibinfo{number}{19} (\bibinfo{year}{2021}).
\newblock
\href{https://doi.org/10.3390/electronics10192329}{doi:\nolinkurl{10.3390/electronics10192329}}


\bibitem[Wang et~al\mbox{.}(2022)]%
        {Wang2022}
\bibfield{author}{\bibinfo{person}{Han Wang}, \bibinfo{person}{Dongdong Wang}, \bibinfo{person}{Haoxiang Liu}, {and} \bibinfo{person}{Gang Tang}.} \bibinfo{year}{2022}\natexlab{}.
\newblock \showarticletitle{A predictive sliding local outlier correction method with adaptive state change rate determining for bearing remaining useful life estimation}.
\newblock \bibinfo{journal}{\emph{Reliability Engineering and System Safety}}  \bibinfo{volume}{225} (\bibinfo{year}{2022}), \bibinfo{pages}{108601}.
\newblock
\showISSN{0951-8320}
\href{https://doi.org/10.1016/j.ress.2022.108601}{doi:\nolinkurl{10.1016/j.ress.2022.108601}}


\bibitem[Yang(1999)]%
        {Yang1999}
\bibfield{author}{\bibinfo{person}{Yiming Yang}.} \bibinfo{year}{1999}\natexlab{}.
\newblock \showarticletitle{An Evaluation of Statistical Approaches to Text Categorization}.
\newblock \bibinfo{journal}{\emph{Information Retrieval}} \bibinfo{volume}{1}, \bibinfo{number}{1-2} (\bibinfo{year}{1999}), \bibinfo{pages}{69--90}.
\newblock
\href{https://doi.org/10.1023/A:1009982220290}{doi:\nolinkurl{10.1023/A:1009982220290}}


\bibitem[Zhao et~al\mbox{.}(2025)]%
        {zhao2025comprehensive}
\bibfield{author}{\bibinfo{person}{Haoru Zhao}, \bibinfo{person}{Baoshan Zhu}, {and} \bibinfo{person}{Boshuang Jiang}.} \bibinfo{year}{2025}\natexlab{}.
\newblock \showarticletitle{Comprehensive assessment and analysis of cavitation scale effects on energy conversion and stability in pumped hydro energy storage units}.
\newblock \bibinfo{journal}{\emph{Energy Conversion and Management}}  \bibinfo{volume}{325} (\bibinfo{year}{2025}), \bibinfo{pages}{119370}.
\newblock


\bibitem[Zonta et~al\mbox{.}(2020)]%
        {ZONTA2020106889}
\bibfield{author}{\bibinfo{person}{Tiago Zonta}, \bibinfo{person}{Cristiano~André {da Costa}}, \bibinfo{person}{Rodrigo {da Rosa Righi}}, \bibinfo{person}{Miromar~José {de Lima}}, \bibinfo{person}{Eduardo~Silveira {da Trindade}}, {and} \bibinfo{person}{Guann~Pyng Li}.} \bibinfo{year}{2020}\natexlab{}.
\newblock \showarticletitle{Predictive maintenance in the Industry 4.0: A systematic literature review}.
\newblock \bibinfo{journal}{\emph{Computers and Industrial Engineering}}  \bibinfo{volume}{150} (\bibinfo{year}{2020}), \bibinfo{pages}{106889}.
\newblock
\showISSN{0360-8352}
\href{https://doi.org/10.1016/j.cie.2020.106889}{doi:\nolinkurl{10.1016/j.cie.2020.106889}}


\end{thebibliography}
